\documentclass[10pt,twocolumn,letterpaper]{article}

\usepackage{cvpr}
\usepackage{times}
\usepackage{epsfig}
\usepackage{graphicx}
\usepackage{amsmath}
\usepackage{amssymb}

\usepackage[breaklinks=true,bookmarks=false]{hyperref}
\hypersetup{
    colorlinks=true,
    linkcolor=blue,
    filecolor=magenta,      
    urlcolor=cyan,
}

\cvprfinalcopy 

\DeclareMathAlphabet{\pazocal}{OMS}{zplm}{m}{n}
\newcommand{\Lb}{\pazocal{L}}
\usepackage{enumitem}
\usepackage{cite}   
\usepackage{enumitem}
\usepackage{graphicx}
\usepackage{subfloat}
\usepackage{tabularx}
\usepackage{adjustbox}
\usepackage{amssymb}
\usepackage{multirow}
\usepackage[flushleft]{threeparttable}
\usepackage{subfigure}

\newcommand{\figref}[1]{Fig.~\ref{#1}}
\newcommand{\tabref}[1]{Table~\ref{#1}}

\renewcommand{\ie}{\textit{i.e.}}
\renewcommand{\eg}{\textit{e.g.}}

\def\cvprPaperID{2089} 

\ifcvprfinal\fi
\begin{document}

\title{Deep Blind Video Decaptioning by Temporal Aggregation and Recurrence}

\author{Dahun Kim\thanks{Both authors contributed equally to this work.} \\
KAIST\\
\and
Sanghyun Woo\footnotemark[1]\\
KAIST\\
\and
Joon-Young Lee\\
Adobe Research\\
\and
In So Kweon\\
KAIST\\
}

\maketitle

\begin{abstract}
  Blind video decaptioning is a problem of automatically removing text overlays and inpainting the occluded parts in videos without any input masks. While recent deep learning based inpainting methods deal with a single image and mostly assume that the positions of the corrupted pixels are known, we aim at automatic text removal in video sequences without mask information. In this paper, we propose a simple yet effective framework for fast blind video decaptioning. We construct an encoder-decoder model, where the encoder takes multiple source frames that can provide visible pixels revealed from the scene dynamics. These hints are aggregated and fed into the decoder. We apply a residual connection from the input frame to the decoder output to enforce our network to focus on the corrupted regions only. Our proposed model was ranked in the first place in the ECCV Chalearn 2018 LAP Inpainting Competition Track2: Video decaptioning. In addition, we further improve this strong model by applying a recurrent feedback. The recurrent feedback not only enforces temporal coherence but also provides strong clues on where the corrupted pixels are. Both qualitative and quantitative experiments demonstrate that our full model produces accurate and temporally consistent video results in real time (50+ fps).
\end{abstract}

\section{Introduction}
Dealing with missing or corrupted data is a crucial step before consuming visual contents. In many applications in image/video processing, such incompleteness degrades the visual perception for both human and machines. To overcome this limitation, recent approaches have focused on solving denoising~\cite{xie2012image}, restoration~\cite{mao2016image}, super-resolution~\cite{dong2016image}, and inpainting~\cite{pathak2016context}. In this paper, we focus on video decaptioning, one of the video inpainting tasks where the solution can be directly applicable to real-world video restoration scenarios.

In the context of media and video data from various languages, there are frequently text captions or encrusted commercials. These text overlays reduce visual attention and occlude parts of frames. Removing the text overlays and inpainting the occluded parts require the understanding of the spatio-temporal context in videos. However, processing a video sequence requires high memory footprint and time complexity due to the additional time dimension.

\begin{figure}[t]
\begin{center}
\begin{tabular}{@{}c@{}}
\includegraphics[width=1.0\linewidth]{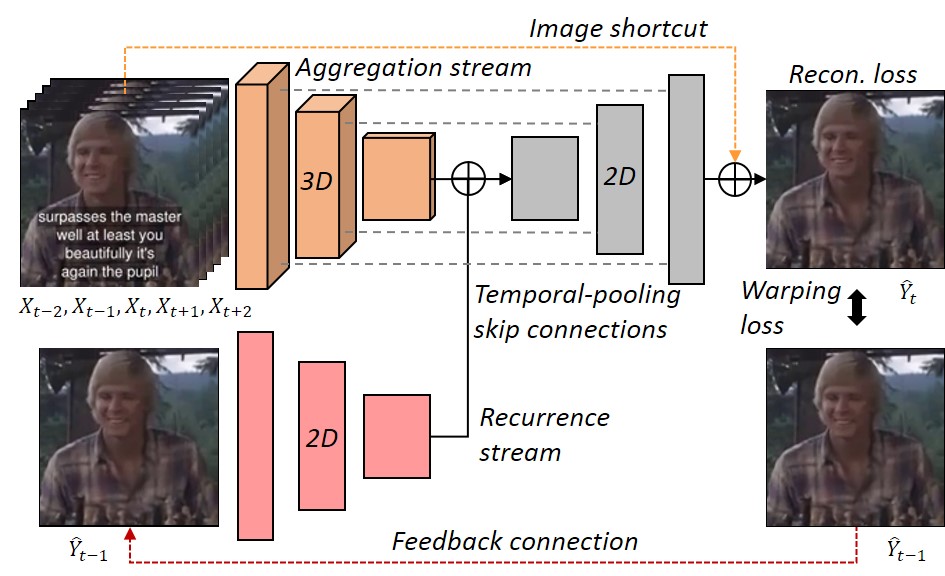} \\
\end{tabular}
\end{center}
\caption{\textbf{Overview of our blind video decaptioning network. (BVDNet)} We propose a hybrid encoder-decoder model, where the aggregation encoder stream takes multiple input frames and the decoder reconstructs the middle frame. The temporal-pooling skip connections carry low-level information. By a residual learning algorithm, our model directly learns to recover the corrupted pixels in the input. The output is then fed into a feedback connection for a recurrent learning to the next time step.}
\label{fig:teaser}
\end{figure}


A straightforward way to perform video decaptioning is to recover a video frame-by-frame. However, it loses a great advantage coming from the video dynamics. In many subtitled videos, the occluded parts in a frame are often revealed in its neighboring frames if the object moves out of the text overlay or the subtitles disappear (\eg~\figref{fig:tc_loss}-a). Also, since dealing with a single frame does not consider any temporal consistency, the consecutive frames in the recovered video are not likely to be connected naturally. The video captions which disappear or change very suddenly and independently to the visual semantics make it more challenging to maintain the temporal stability. Post-processing with off-the-shelf techniques like blind video temporal consistency~\cite{lai2018learning} is also not applicable, since they require reference video sequences with dense optical flows which are not reliable in our case due to the corrupted regions. 

Another challenge in automatic text removal is that the binary indicator (\ie~inpainting mask) for the corrupted pixels is not given in advance. In contrast, most existing inpainting methods~\cite{pathak2016context,yang2017high,yu2018generative} usually assume that the binary pixel mask is available and use different image priors based on it. We cannot directly adopt their scenarios because annotating (or creating) such pixel masks for every frame in videos is impractical and limits the system's autonomy. Furthermore, many video subtitles include semi-transparent shadows (as in \figref{fig:teaser}, \figref{fig:tc_loss}-a, b) where it is ambiguous to label the pixels in binary. These regions should not be considered as solid occlusions, because of the underlying information.

To overcome the aforementioned challenges in blind video decaptioning, we propose a simple yet effective encoder-decoder model (see \figref{fig:teaser}), where the encoder aggregates spatio-temporal context from neighboring source frames and the decoder reconstructs the target frame. We apply a residual learning algorithm where our network is encouraged to touch only the corrupted pixels. We further introduce a recurrent feedback connection, so that the generation of the target frame is based on the previously generated frame. Since the features from the neighbor frames and the previous output are largely dissimilar on the corrupted regions, it helps our network to better detect corrupted pixels and boosts the performance. We train our model with the gradient reconstruction loss and the structural similarity loss in conjunction with the conventional L1 loss. We validate the contribution of our design components through experiments. To our best knowledge, this is the first attempt to apply deep learning to the blind video inpainting application.

Our contribution can be summarized as follows:
\begin{itemize}[topsep=1pt]
\item Unlike most of the existing methods for image/video inpainting, the proposed approach aggregates neighboring spatio-temporal features in the encoder and recovers a decaptioned frame in the decoder without requiring the inpainting masks.
\item We design an effective and robust loss function for video decaptioning and empirically validate the usefulness of the loss terms with our architectural design by extensive ablation study. 
\item Our model outperforms other competing methods and runs in real time (50+ fps). We took the first place in the ECCV Chalearn 2018 LAP Video Decaptioning Challenge.
\item We further improve our model by introducing a recurrence mechanism and boost the performance even more in terms of both visual quality and temporal coherency. 
\end{itemize}

\section{Related Work}
\label{sec:related}

\subsection{Image Inpainting}
Approaches of traditional image inpainting make use of the image-level features to diffuse the texture from the surrounding context to the missing hole~\cite{ballester2001filling,bertalmio2000image}. These methods can only tackle small holes and would lead to noise patterns and artifacts for large holes. Later works using patch-based methods could optimize the inpainting performance by searching the best matching patches~\cite{efros1999texture,barnes2009patchmatch}. However, while these methods could provide plausible texture generation in the hole, they are not aware of the high-level semantics of the image and cannot make reasonable inference for object completion and structure prediction.

Recently, many image inpainting models based on Convolutional Neural Networks (CNNs) have been proposed~\cite{pathak2016context,yang2017high,yeh2017semantic,iizuka2017globally,yu2018generative,liu2018image,yu2018free}. These approaches directly infer pixel values inside holes in an end-to-end fashion. Thanks to their ability to learn adaptive image features of various semantics, they can synthesize pixels that are more visually plausible. Pathak~\etal~\cite{pathak2016context} introduced Context Encoder that enabled to fill large masks using a CNN model. They proposed to use Generative Adversarial Networks (GAN)~\cite{goodfellow2014generative} to avoid the blurring artifacts. Iizuka~\etal~\cite{iizuka2017globally} proposed a fully convolutional network with both global and local discriminators to obtain semantically and locally coherent image inpainting results. This method, however, heavily relies on a post-processing step; Poisson image blending. Yu~\etal~\cite{yu2018generative} proposed a coarse-to-fine model by stacking two generative networks, ensuring the color and texture consistency of generated regions with surroundings. Moreover, in order to capture long-range spatial dependencies, a contextual attention module is integrated into the networks. However, this model does not generalize well on irregular masks because it is mainly trained on large rectangular masks. To better handle free-form masks, partial convolution~\cite{liu2018image} and gated convolution~\cite{yu2018free} are proposed where the convolution weights are masked or re-weighted respectively to utilize valid pixels only.

\begin{figure}
\includegraphics[width=1.0\linewidth]{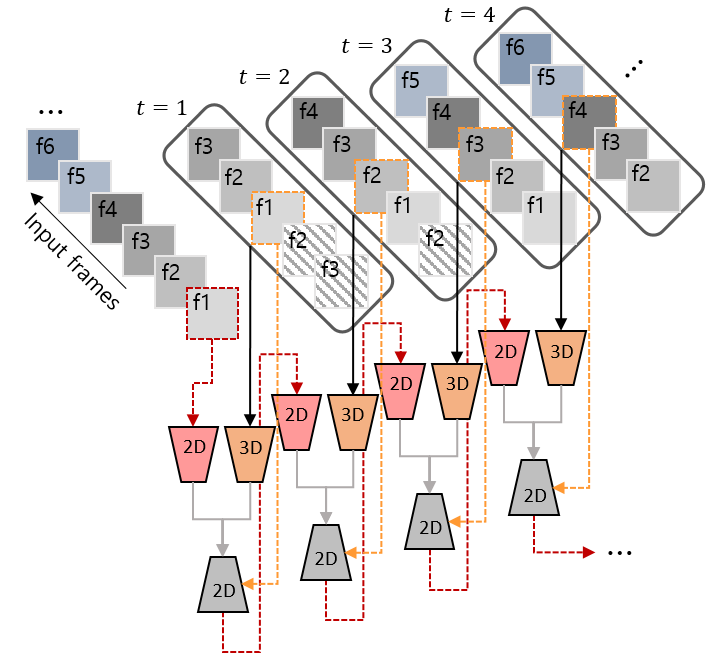}
\caption{\textbf{Multiple time steps of our BVDNet framework.} The image shortcuts, input to the recurrence stream, and input to the aggregation stream are represented by orange, red, and black arrows, respectively. We mirror-pad at initial input boundary (\eg  $t$=1,2) as illustrated in stripe pattern. Our method processes frames sequentially in a sliding window manner and produces the video results in real time.}
\label{fig:bvdnet_inference}
\end{figure}

\subsection{Video Inpainting}

Approaches for video inpainting can be categorized into object-based and patch-based methods. Object-based approahces~\cite{cheung2006efficient,jia2006video,ling2011virtual} segment a video into moving foreground objects and background that is either still or exhibits smooth motion. The moving objects are, in general, copied into the holes as smoothly as possible, whereas the background is inpainted using image inpainting methods. However, such approaches work well only when the objects' motion shows strict periodicity.

The patch-based method~\cite{patwardhan2005video} copied and pasted small video patches into the holes. Patwardhan~\etal~\cite{patwardhan2007video} further improved this approach so that moving cameras could be dealt with. Granados~\etal~\cite{granados2012not} proposed a semi-automatic algorithm which optimizes the spatio-temporal shift map~\cite{pritch2009shift}. However, manual tracking of moving objects is required to reduce a large search space and high computational complexity. Further, Wexler~\etal~\cite{wexler2004space} proposed an iterative method to solve a global optimization method. However, these methods rely on the inpainting masks for each frame and lack high-level semantic understanding. Newson~\etal~\cite{newson2014video} extend this by developing a 3D version of PatchMatch~\cite{barnes2009patchmatch}. Huang~\etal~\cite{huang2016temporally} propose to use additional optical flow term by modifying the energy term of ~\cite{wexler2004space} for temporally coherent predictions in results.

Our work takes an important next step beyond the aforementioned prior works in that we address a blind video inpainting task using a deep CNN model. We propose a 3D encoder-2D decoder model that can effectively learn spatio-temporal features to recover clear frames in a data-driven manner. Our predicted frames are visually natural and temporally smooth without any post-processing step. Thanks to our light-weight design, the overall process performs in real-time (50+ fps).

\section{Proposed Method}
\label{sec:method}

Video decaptioning aims to estimate original frames \{$\hat{Y}$\} from the subtitled, noised frames \{$X$\}. The recovered region should either be as accurate as in the GT frames \{$Y$\} or seamlessly merged into the surrounding pixels. Our strategy is to collect hints from the multiple neighboring (source) frames and recover a target frame. This is to leverage the scene dynamics in a video where the occluded parts are often revealed in the lagging or leading frames as the objects move or the subtitles change. We also propose to use a recurrent feedback connection as an additional source stream. This helps our network to reduce temporal flickering and to automatically detect the corrupted regions. 
 
\subsection{Residual Learning}
Directly estimating all pixels in a frame may needlessly touch uncorrupted pixels. To deal with the absence of the pixel indicators (inpainting masks), we train our model by a residual learning algorithm. Specifically, the final output is yielded by summing the input center frame \{$X_t$\} and the predicted residual image \{$R_t$\} in a pixel-wise manner. This encourages our network to explicitly focus on the corrupted pixels only, and also prevent the global tone distortion.

Formally, with the proposed decaption model $f$, we model the video captioning problem as 
\begin{eqnarray}
\hat{Y_t} = f(X_{t-N:t+N}, \hat{Y}_{t-1}) + X_t,
\label{equ:decaptioning_eq}
\end{eqnarray}
where $t$ denotes a frame index and $N$ is a temporal radius. 

\subsection{Network Design}
The overall decaptioning algorithm is illustrated in \figref{fig:teaser}. Our core design is a hybrid encoder-decoder model, where the encoder consists of two sub-networks: 3D CNN and 2D CNN. The decoder follows a normal 2D CNN design as in other image generation networks. The network is designed to be fully convolutional, which can handle arbitrary size input. The final output video is obtained by applying $f$ in an auto-regressive manner as in \figref{fig:bvdnet_inference}.

\noindent \textbf{Two-stream encoder.} \quad
Our strategy is to collect potential hints from multiple source frames that can provide visible pixels revealed from the scene dynamics. Also, we enforce the generation of the target frame to be consistent with the previous generation. We construct a two-stream hybrid encoder where each source stream is trained to achieve our objectives. The first encoder stream consists in 3D convolutions which can directly capture spatio-temporal features from the neighboring frames. This can help in understanding the short-term video-level context which is required to recover the target frame. The input tensor shape is $H\times W\times T\times C$, where $H$,$W$ and $C$ are the height, width and channels of the input frame $\{X\}$, and $T = 2N+1$. We use $N=2$ in our network ($T=5$). Here, the goal is to remove text overlays in the center input frame (3th out of 5). The temporal dimension gradually reduces into 1 passing through the 3D convolution layers in this stream.

The second stream is a 2D CNN which takes the previously generated frame, of size  $H\times W\times 1\times C$, as input. This stream provides a reference for the current generation to be temporally coherent with. Moreover, the encoded feature is combined with the \textit{temporally-pooled one-frame} feature from the first stream by element-wise summation. Since the features from the two streams are comparatively different on the corrupted regions, the combined hybrid feature map implicitly encodes the knowledge on where to attend.

\noindent \textbf{Bottleneck and temporal-pooling skip connections.}

The encoder is followed by bottleneck layers that consist of several dilated convolutions, as suggested in~\cite{iizuka2017globally}. The large receptive field size helps to capture wide spatial context which supports the recovery of the corrupted pixels. The following is a 2D CNN decoder which is symmetric to the 2D encoder stream. 

We apply skip connections only between the 3D encoder stream and the decoder. Each skip connections pass through a 3D convolution layer that pools the temporal dimension into \textit{one frame}, so that the feature map can be directly concatenated with the decoder features of equal dimension. Despite the concern raised by Yu~\etal\cite{yu2018free} that the skip connections carry almost zero features on the corrupted regions, our temporal-pooling skip connections are immune to this problem since they can adaptively aggregate low-level features that are complementary to the occluded points in the center frame. 

Our full network is trained to generate the residual frame \{$R_t$\}, which is added with the input center frame \{$X_t$\} to produce the final output \{$\hat{Y_t}$\}. %

\subsection{Frame Sampling}
As we mentioned earlier, our task can greatly benefit from the video dynamics. If the scene moves or the subtitles disappear in the neighboring frames, the occluded parts will be revealed, which provides critical clues to the underlying content. To maximize this gain, we attempt to find the optimal frame sampling interval for our model. With the minimum interval of 1, the input frames will contain non-significant dynamics. If we jump with large stride, on the other hand, irrelevant new scenes will be included. We empirically find that the stride of 3 performs the best in our preliminary experiment. Since we use $T=5$, our model has about 15 frame-term view range.

\subsection{Loss Functions}
\label{section:loss_functions}
We train the sequential video decaptioning model $f$ by solving the following objective function,
\begin{eqnarray}
\min_{f} \left( \lambda_{R}\mathcal{L}_{R}(f) + \lambda_{T}\mathcal{L}_{T}(f) \right),
\end{eqnarray}
where $\mathcal{L}_{R}$ is an \textbf{image reconstruction loss}, and $\mathcal{L}_{T}$ is a \textbf{temporal consistency loss}. $\lambda_{R}$ and $\lambda_{T}$ denote to the weighting coefficients which are set to 1 and 2 throughout the experiments.

To address image reconstruction, a simple way is to minimize the L1 loss following the previous studies~\cite{liu2018image,yu2018generative}. For the structural details, We apply the SSIM loss~\cite{wang2004image} with a small patch window according to the setting of the competition evaluation metric. Inspired by~\cite{eigen2015predicting}, we also use a first-order matching term, which compares image gradients of the prediction with the ground truth, and encourages the prediction to have not only close-by values but also similar local structure. To this end, the \textbf{image reconstruction loss} $\mathcal{L}_{R}$ includes three terms as 
\begin{eqnarray}
\Lb_{1} = \left \|  \hat{Y_t} - Y_t \right\|_{1}, \\ 
\Lb_{SSIM} =  (\frac{{(2\mu_{\hat{Y_t}} \mu_{Y_t} + c_1 )(2\sigma_{\hat{Y_t} {Y_t}} + c_2)}}
{{(\mu_{\hat{Y_t}}^2+\mu_{Y_t}^2 +c_1)(\sigma_{\hat{Y_t}}^2+\sigma_{Y_t}^2 +c_2)}}), \\
\Lb_{grad.} = \left \|\nabla_{W}(\hat{Y_t} - Y_t)\right\|_{1} + \left \|\nabla_{H}(\hat{Y_t} - Y_t)\right\|_{1}, \\
\Lb_{R} = \Lb_{1} + \Lb_{SSIM} + \Lb_{grad.},
\label{eqn:loss_R}
\end{eqnarray}
where $\hat{Y_t}, Y_t$ denote the predicted and target groundtruth frames respectively. $\mu, \sigma$ denote the average, variance. $c_1, c_2$ denote two stabilization constants which are respectively set to $0.01^2, 0.03^2$. $\nabla_{W}, \nabla_{H}$ are the image gradients along the horizontal and vertical axis.

With the recurrence (second) stream in the encoder, we optimize our model with additional temporal warping loss which is widely used in video generation works~\cite{lai2018learning,wang2018video,huang2017real}. 
The \textbf{temporal consistency loss} $\mathcal{L}_{T}$ is defined as
\begin{eqnarray}
\Lb_{T} = \sum\limits_{t=1}^{T-1} M_{t-1}^{t} \left \| \hat{Y_t} - \phi({Y}_{t-1})\right\|_{1},
\end{eqnarray}
where $M$ represents the binary occlusion mask and $\phi$ denotes the flow warping operation. We use optical flow between consecutive target frames obtained by FlowNet2~\cite{ilg2017flownet}, to compute our temporal loss. For the training, we set the number of recurrences to 5 $(T=5)$.

\begin{table*}
\centering
\setlength{\tabcolsep}{5pt}
\begin{tabular}{c | c c c | c c c | c | c c c }
\hline
&\multicolumn{3}{c |}{Architecture}  &\multicolumn{3}{c |}{Losses} &\multicolumn{1}{c |}{Recurrence}  &\multicolumn{3}{c}{Evaluation Metric}\\
Exp  &3D-3D &2D-2D & 3D-2D & L1 & grad. L1 & SSIM  & Enc. Stream & MSE & PSNR & DSSIM \\
\hline
\hline    
1  & \checkmark &  &  & \checkmark &    &    &    &0.0031 & 28.4590 & 0.0652 \\
2  &   & \checkmark & & \checkmark &    &    &    &0.0012 & 33.6803 & 0.0279 \\
3  &   & & \checkmark & \checkmark &    &    &    &0.0011 & 34.1029 & 0.0261 \\
4  &   &  &\checkmark  & \checkmark & \checkmark &    &    &0.0010&34.2251&0.0276	 \\
5  &   &  &\checkmark  & \checkmark & \checkmark &\checkmark &    &0.0010&34.6544&0.0225 \\
\hline
\hline
6 (\textbf{Our full model})  &   &  &\checkmark  & \checkmark & \checkmark &\checkmark &\checkmark  &\textbf{0.0010}&\textbf{34.7055}&\textbf{0.0222} \\
\hline
\end{tabular}
\caption{The ablation studies on architectural design, loss functions, and recurrence stream. We evaluate on ChaLearn 2018 LAP Inpainting Track2 \textit{validation} set. }
\label{tab:ablation}
\end{table*}

\section{Implementation}
\label{sec:impl}
\noindent \textbf{Dataset.} \quad We used the ECCV Chalearn 2018 LAP Video Decaptioning Challenge dataset for training, validation, and testing. It is a large dataset of 5 seconds (125 frames) MP4 video clips in $128 \times 128$ pixel RGB frames, containing both encrusted subtitles (\{$X$\}) and without subtitles (\{$Y$\}). The dataset contains a wide variety of captions with different colors, size, positions, and shadows. The training and validation set consist in 70K and 5K sample pairs of input and ground truth video clips, respectively. The testing set consists of 5K input video clips without ground truth. We convert every video clip into PNG images in our experiments.

\noindent \textbf{Training.} \quad 
We adopt horizontal flipping and color jittering for data augmentation. We train our model for 200 epochs with a batch size of 128. Adam optimizer is used with $\beta$ = (0.9, 0.999) and a learning rate of 0.001. The training takes 3 days on two NVIDIA GTX 1080 Ti GPUs. For the competition, we train our model without the recurrence stream in the encoder and the warping loss.

\noindent \textbf{Testing.} \quad For the pixels where the absolute difference between the input middle frame $\{X_t\}$ and the prediction $\{\hat{Y_t}\}$ is less than 0.01 in $[0,1]$ scale, we copy the values from the input frame. Finally, we convert PNG files back to MP4 videos.

\noindent \textbf{Evaluation Metric.} \quad To evaluate the quality of the reconstruction, the mean square error (MSE), the peak signal-to-noise ratio (PSNR), and the structural dissimilarity (DSSIM -- \ie~(1-SSIM)/2) are used.

\begin{figure*}
\begin{center}
\def\arraystretch{1.0}
\small
\begin{tabular}{@{}c@{\hskip 0.01\linewidth}c@{\hskip 0.01\linewidth}c@{\hskip 0.01\linewidth}c@{\hskip 0.01\linewidth}c@{}}

    \includegraphics[width=0.18\linewidth]{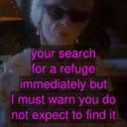}&
    \includegraphics[width=0.18\linewidth]{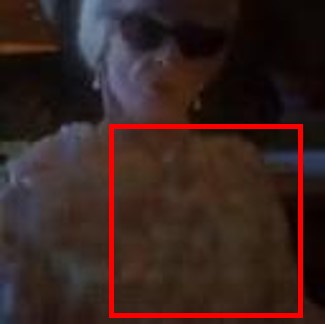}&
    \includegraphics[width=0.18\linewidth]{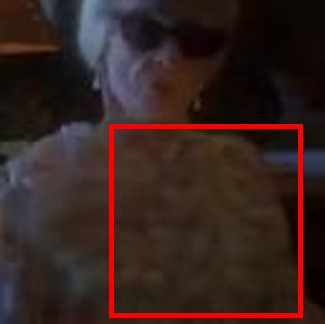}&
    \includegraphics[width=0.18\linewidth]{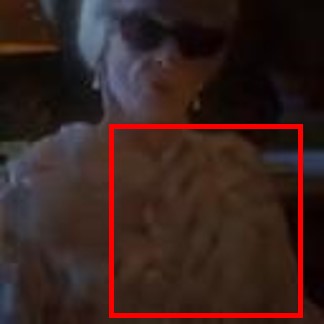}&
    \includegraphics[width=0.18\linewidth]{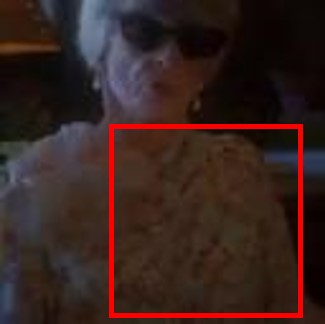}\\

    \includegraphics[width=0.18\linewidth]{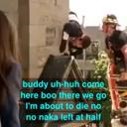}&
    \includegraphics[width=0.18\linewidth]{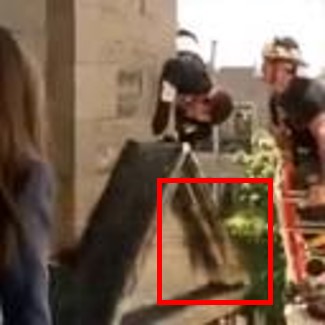}&
    \includegraphics[width=0.18\linewidth]{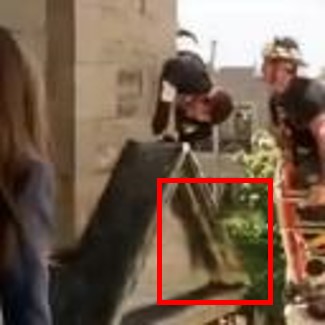}&
    \includegraphics[width=0.18\linewidth]{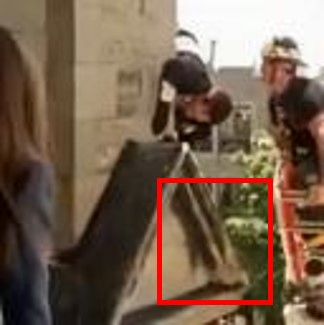}&
    \includegraphics[width=0.18\linewidth]{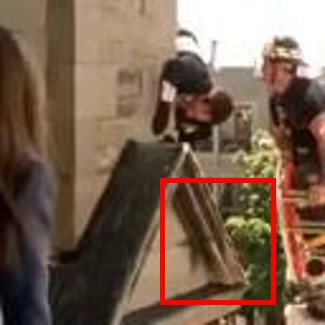}\\
    
{(a) input} & {(b) L1} & {(c) L1 + grad.L1} & {(d) L1+grad.L1+SSIM} & {(e) target (groundtruth)} \\

\end{tabular}
\end{center}
\caption{\textbf{The impact of each loss terms.} (a) An input center frame. (b-d) The reconstructed frames with: (b) L1 loss, (c) L1 +
gradient L1 loss, and (d) L1 + gradient L1 + SSIM loss. (e) Ground truth frame. \textit{Best viewed when zoomed-in}.}
\label{fig:losses}
\end{figure*}

\begin{figure*}
\begin{center}
\def\arraystretch{1.0}
\small
\begin{tabular}{@{}c@{\hskip 0.01\linewidth}c@{}}
    \includegraphics[width=0.49\linewidth]{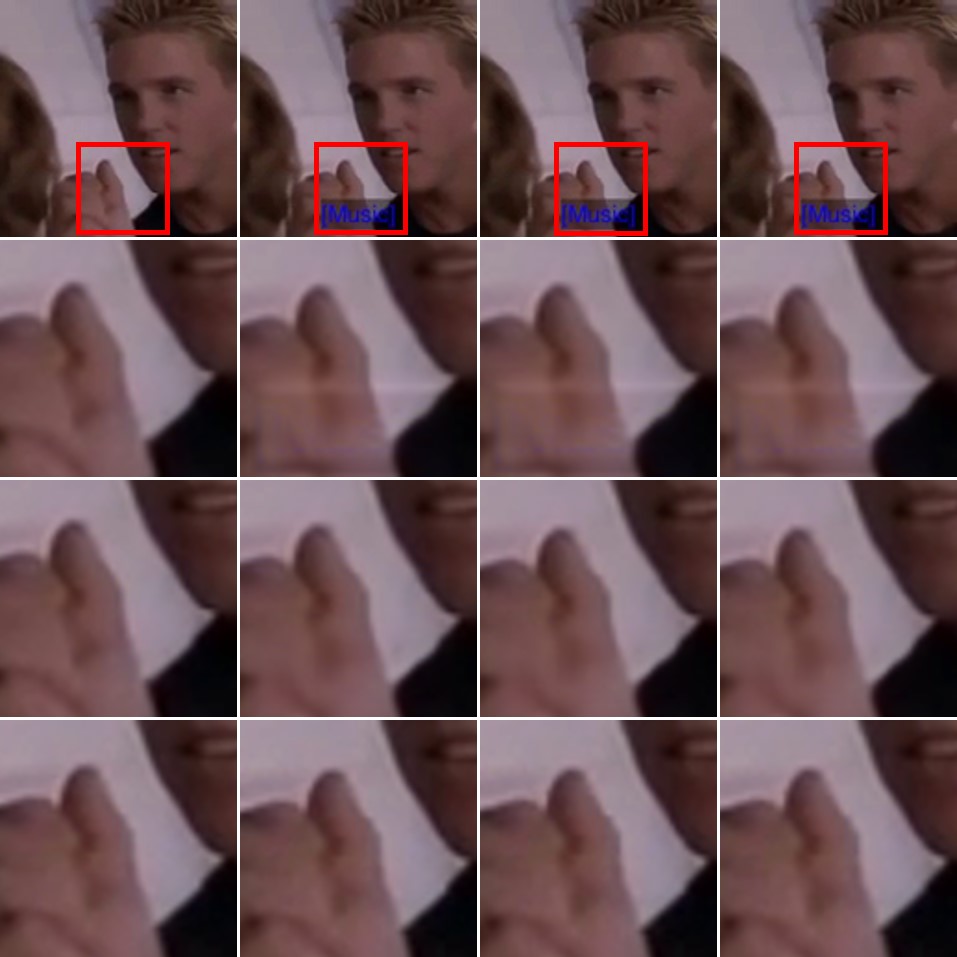}&
    \includegraphics[width=0.49\linewidth]{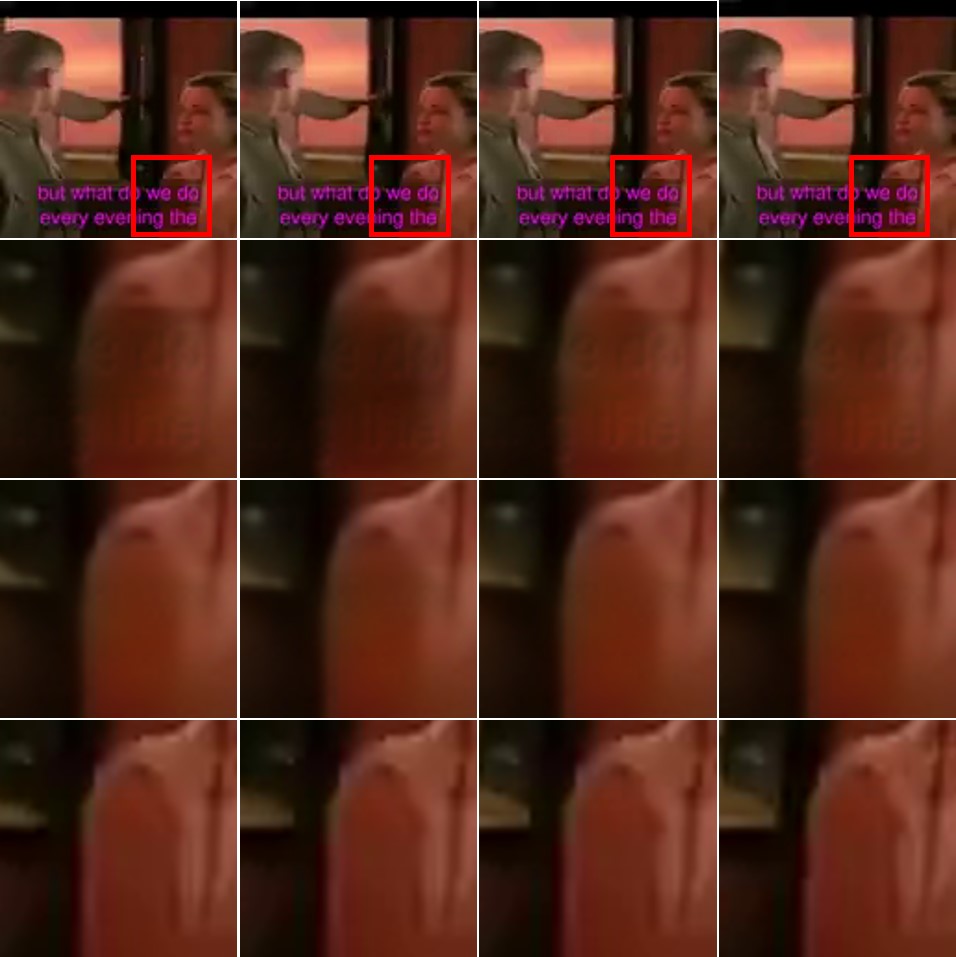} \\    
{(a)} & {(b)} \\
\end{tabular}
\end{center}
\caption{\textbf{The impact of recurrence on temporal consistency.} For each sample, we visualize four consecutive input frames in the top row. In the bottom rows are the zoomed-in views of our results \textit{without recurrence} (2nd row), \textit{with recurrence} (3rd row), and the ground truth frames (4th row). Without the recurrence, the change in the subtitles leads to temporally flickering artifacts (a)}
\label{fig:tc_loss}
\end{figure*}

\begin{table}
\centering
\setlength{\tabcolsep}{10pt}
\begin{tabular}{ c | c | c | c }
\hline
	   & Ours	  	& + GAN loss & - Skip\\
\hline
MSE   & \textbf{0.0010} 	    & 0.0015	 & 0.0010\\
PSNR  & \textbf{34.7055}     	& 31.2257    & 34.3892\\
DSSIM & \textbf{0.0222}	 		& 0.0384     & 0.0233\\
\hline
\end{tabular}
\caption{The ablation studies on additional GAN loss and without residual learning. We evaluate on ChaLearn 2018 LAP Inpainting Track2 \textit{validation} set. }
\label{tab:ablation2}
\end{table}

\begin{table}
\centering
\setlength{\tabcolsep}{8pt}
\begin{tabular}{ c|c|c|c|c }
\hline
& Value & MSE & PSNR & DSSIM \\
\hline
              &3         	&  0.0011  & 33.7895 & 0.0247 \\
Number        &\textbf{5}    &  \textbf{0.0010}  & \textbf{34.7055} & \textbf{0.0222} \\
of frames     &7             &  0.0010  & 34.5063 & 0.0229 \\
              &9             &  0.0010  & 34.6260 & 0.0226 \\
\hline
\end{tabular}
\caption{The ablation studies on the hyperparamter: \textit{number of input frames}. We evaluate on ChaLearn 2018 LAP Inpainting Track2 \textit{validation} set. }
\label{tab:hyperparam}
\end{table}

\begin{table}
\centering
\setlength{\tabcolsep}{5pt}
\begin{tabular}{ c|c }
\hline
Encoder version & Temporal Errors \\
\hline
\textit{without} recurrence (Ours-Exp 5) &  0.00117 \\
\textit{with} recurrence (Ours-Exp 6)   &  \textbf{0.00090} \\
\hline
\end{tabular}
\caption{Temporal errors (warping errors) of our full model with and without temporal consistency constraints. We evaluate on 500 clips of ChaLearn 2018 LAP Inpainting Track2 \textit{validation} set. }
\label{tab:temporal}
\end{table}

\section{Experimental Results}
\label{sec:results}

\subsection{Ablation Study}
In order to evaluate the effectiveness of different components of the proposed BVDNet, we conduct ablation studies using the publicly released validation set.

\noindent \textbf{The impact of 3D encoder stream.}\quad  One of our core design choices is to use a 3D CNN encoder stream in conjunction with a following 2D decoder. To validate the effectiveness of this design, we construct two naive baselines to compare with: a 3D encoder-3D decoder and a 2D encoder-2D decoder models. We note that all models in this experiment contain a single-stream encoder without the recurrence stream encoder for a clearer comparison. We construct all the models with a comparable number of parameters. As shown in Exp 1, 2, and 3 in \tabref{tab:ablation}, our 3D-2D model shows the best performances in all three metrics. This implies that spatio-temporal feature extraction from the neighboring frames indeed helps our target task, providing our model with a distinct advantage over the \textit{frame-by-frame} competitor. On the other hand, it is empirically shown that adopting heavy 3D-3D operations does more harm than good. This implies that making use of the neighboring frames does not always work, rather the careful architectural design is required.

\noindent \textbf{The impact of loss functions.} \quad
We test our loss functions both quantitatively and qualitatively. First, we remove each loss terms gradually from our full loss function. Again, we use models with the single-stream encoder, and thus the temporal warping loss is not considered in this experiment. As shown in Exp 3, 4, and 5 in \tabref{tab:ablation}, the best scores are obtained in all metrics when all L1, gradient L1, and SSIM losses are used together. 

We also provide qualitative analysis as shown in~\figref{fig:losses}. The model trained with the L1 loss alone produces relatively blurry outputs. We alleviated this problem by adding the gradient L1 loss and the SSIM loss. We attempt to use the adversarial loss, but it tends to decrease the performance on the evaluation metrics (see ~\tabref{tab:ablation2}). We observe that the gradient L1 and SSIM losses together help recover finer structures (texture and edge) and achieve better evaluation scores as well. In short, the L1 loss plays a role of capturing the overall structure of the corrupted region. The gradient L1 loss and the SSIM loss reduce the artifacts, encouraging the preservation of the local structures.

\noindent \textbf{The impact of the recurrence encoder stream.} \quad
We investigate the effectiveness of our recurrence stream in the encoder, together with the temporal warping loss. We evaluate both frame-level image quality and temporal consistency. As shown in Exp 5 and 6 in \tabref{tab:ablation}, the recurrence stream improves the visual quality of the video results. In addition, we quantitatively compare the temporal consistency of our models with and without the recurrence stream. We measure the temporal error over a video sequence, which is the average pixel-wise Euclidean color difference between consecutive frames. We use FlowNet2~\cite{ilg2017flownet} to obtain pseudo-groundtruth optical flows as in the training. \tabref{tab:temporal} shows that the temporal error is significantly reduced by having the recurrence stream in the encoder. Our approach does not sacrifice either visual quality and temporal stability. This is also shown qualitatively in \figref{fig:tc_loss}. These results imply that the recurrence stream helps our model to better detect and inpaint the corrupted regions automatically, in a temporally coherent manner.



\noindent \textbf{Adding GAN loss.} \quad
The adversarial training encourages the decaptioning results to move towards the natural image manifold. We test the effect of the adversarial training by adding the GAN loss on top of our full loss function. We use $8 \times 8$ PatchGAN~\cite{isola2017image} as our discriminator network that aims to classify whether $8 \times 8$ overlapping image patches are real or fake. However, we observe no visible qualitative improvement and the quantitative performance slightly dropped (\tabref{tab:ablation2}), which is consistent with the results in~\cite{wang2018perceptual}. 

\noindent \textbf{Removing residual image shortcut.} \quad We investigate the importance of the residual learning. If we remove the skip connection from the input center frame to the decoder output, the network should predict the uncorrupted output without referencing the input pixels. As shown in the ~\tabref{tab:ablation2}, adopting the residual learning scheme shows better performances, demonstrating that devoting to the pixels to be recovered is more effective for video decaptioning.

\noindent \textbf{Number of input frames.} \quad In \tabref{tab:hyperparam}, we perform an experiment to determine the hyperparameter $T$ for our model, which is the number of input frames. The \textit{number of input frames} directly relates to the size of input batches, which enables to control the amount of temporal information to be considered at once.
\tabref{tab:hyperparam} shows the comparison results with four different input frame values. We observe that the performance tends to be good with larger input frames in general, while the value of 5 gives the best results. This indicates that having proper temporal view range is crucial for video decaptioning.

\subsection{Model Inference Time}
Our full model has a total of 23 layers and 10.5M parameters. Our model is implemented on Pytorch v0.3, CUDNN v6.0, CUDA v8.0,
and run on the hardware with Intel(R) Xeon(R) (2.10GHz) CPU and NVIDIA GTX 1080 Ti GPU. The model runs at 62.5 fps on a GPU for frames of resolution 128 $\times$ 128 px.

\begin{table}
\centering
\setlength{\tabcolsep}{10pt}
\begin{tabular}{  l | c | c | c  }
\hline
	  & MSE & PSNR & DSSIM \\
\hline
stephane & 0.0022 & 30.1856 & 0.0613 \\
hcilab  & 0.0012 & 33.0228 & 0.0424  \\
anubhap93  & 0.0012 & 32.0021 & 0.0499 \\
arnavkj95  & 0.0012 & 32.1713 & 0.0482 \\
\hline \hline
Ours  & \textbf{0.0011} & \textbf{33.3527} & \textbf{0.0404} \\
\hline
\end{tabular}

\caption{Final performances of the top entries in the ECCV ChaLearn 2018 LAP Inpainting Challenge Track2 \textbf{test phase}. We note that stephane's is the baseline from the organizers~\cite{chalearn}.}
\label{tab:scores}
\end{table}

\subsection{Final Challenge Results}
\noindent \textbf{Quantitative results.} 
\tabref{tab:scores} summarizes the top entries from the leaderboard of ECCV ChaLearn 2018 Inpainting Challenge Track2. We participated with our model \textit{without} the recurrence stream and achieved the first place on the final test phase. The source code and factsheet are publicly available at \url{http://chalearnlap.cvc.uab.es/challenge/26/track/31/result/fact-sheet/237/}.

Our full model is even stronger on the validation set as shown in \tabref{tab:temporal}, but we cannot evaluate the full model on the testing set because the test server is closed.

\begin{figure*}
\begin{center}
\def\arraystretch{1.0}
\small 
\begin{tabular}{@{\hskip 0.009\linewidth}c@{\hskip 0.012\linewidth}c@{\hskip 0.009\linewidth}}
    \includegraphics[height=0.15\linewidth]{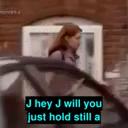}&
    \includegraphics[height=0.15\linewidth]{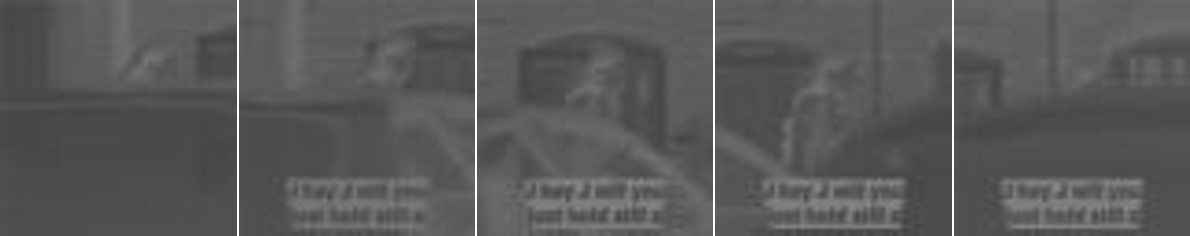}\\
    (a) input & (b) enc-64 \\
\end{tabular}
\begin{tabular}{@{\hskip 0.009\linewidth}c@{\hskip 0.008\linewidth}c@{\hskip 0.008\linewidth}c@{\hskip 0.008\linewidth}c@{\hskip 0.009\linewidth}}
    \includegraphics[height=0.15\linewidth]{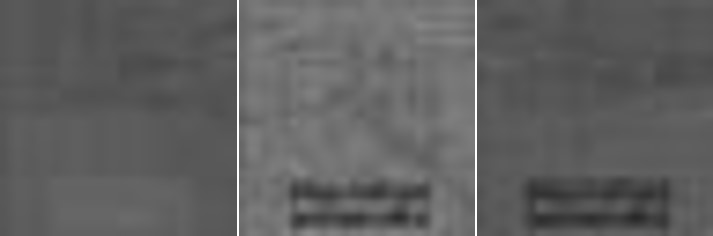}&
    \includegraphics[height=0.15\linewidth]{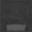}&
    \includegraphics[height=0.15\linewidth]{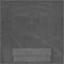}&
    \includegraphics[height=0.15\linewidth]{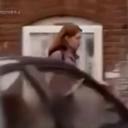}\\
    (c) enc-32 & (d) dec-32 & (e) dec-64 & (f) output
\end{tabular}

\end{center}
\caption{\textbf{Visualization of learned feature activation.} For the visualization, we average each feature maps along the channel axis, perform zero-one normalization, and up-sample to $128 \times 128$ px. The numbers in the labels denote spatial resolution of the feature maps. We observe hierarchical attention operations across the network. In the early encoder layers (b, c), low-level features such as background textures (~\eg \textit{around the subtitles}) are aggregated along the time dimension. The latter decoder layers (d, e) then gradually focus on the exact target regions (~\eg \textit{on the subtitles}) which require high-level semantics to be synthesized. }
\label{fig:gate_vis}
\end{figure*}

\noindent \textbf{Qualitative results.} \quad We visualize the learned feature maps of our full model in \figref{fig:gate_vis}. We observe a hierarchical attention where the 3D encoder layers captures low-level features such as background texture features along time axis, and the 2D decoder layers then gradually \textit{attend to} the exact corrupted region to recover the original content. 

\begin{figure*}
\begin{center}
\def\arraystretch{1.0}
\small 
\begin{tabular}{@{}c@{\hskip 0.01\linewidth}c@{}}
    \includegraphics[width=0.49\linewidth]{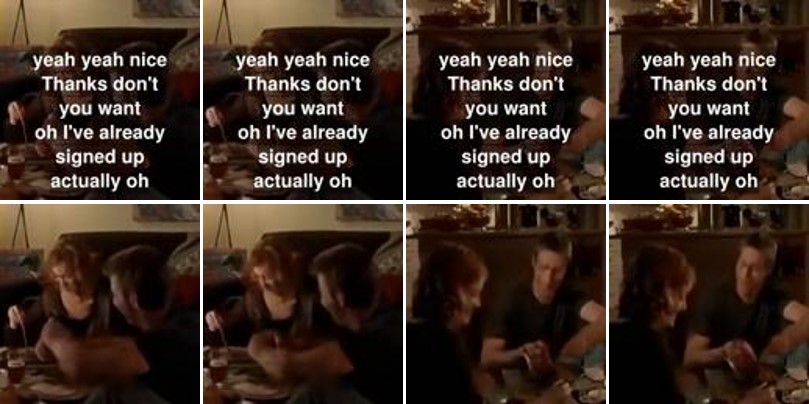}&
    \includegraphics[width=0.49\linewidth]{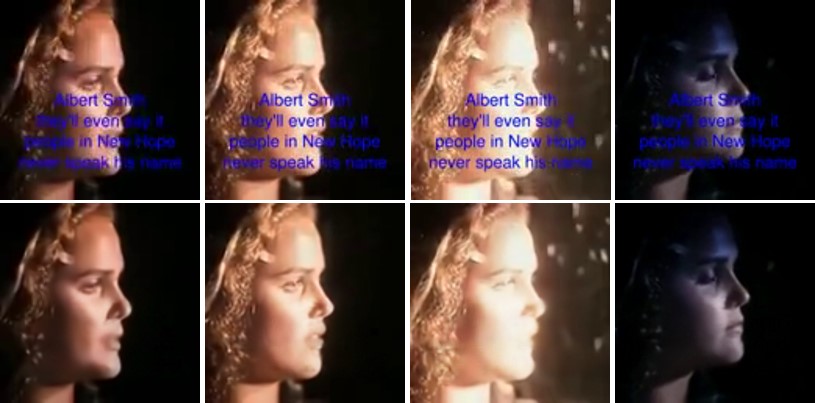}\\ 
    (a) & (b) \\
    \includegraphics[width=0.49\linewidth]{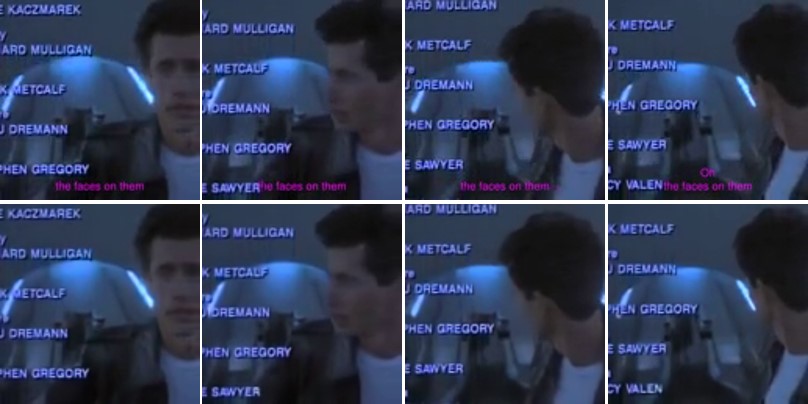}&
    \includegraphics[width=0.49\linewidth]{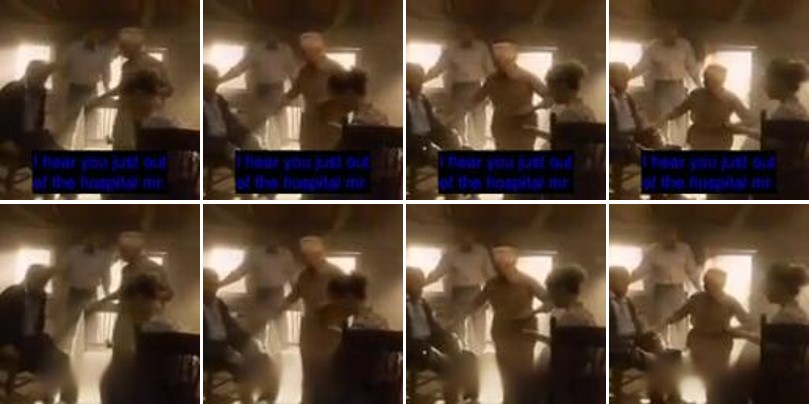}\\
    (c) & (d)
\end{tabular}
\end{center}
\caption{\textbf{Qualitative decaptioning results.} For each example, the top rows are the input sequences and the bottom rows are the decaptioning results using our full model. For visualization, we determine the time interval between the frames to be 0.1 seconds. Our model performs well on various types of subtitles with complex background variations and also is able to separate the non-caption texts in a video.}
\label{fig:qual_vis}
\end{figure*}

\figref{fig:qual_vis} shows examples of our decaptioning results. Our full model successfully recovers the video frames with smooth temporal transition even when there are active object movements as in~\figref{fig:qual_vis}-(a). Also, fine details and textures are well reconstructed even when heavy illumination change exists as in~\figref{fig:qual_vis}-(b). Even when there are non-caption texts in the video, \eg~\figref{fig:qual_vis}-(c), our algorithm is able to separate between the text overlays and texts coming from the video. However, the results are relatively blurry when the input frames have a solid shadow which makes complete occlusions, as in~\figref{fig:qual_vis}-(d). This is probably due to the lack of samples with such solid shadows in the given training set. 

\section{Conclusion}
\label{sec:conclusion}
In this paper, we propose a deep network model for fast blind video decaptioning that learns to remove text overlays in videos. Our model collects hints from not only the current frame but also the future and the past neighboring frames. In addition, it generates each frame conditionally to the previous output frame for the temporal consistency preserving. We design an encoder-decoder model, where the hybrid encoder consists of a 3D CNN stream and a recurrent feedback stream. The spatio-temporal features from these multiple source streams are extracted and fed into the image-based decoder. The skip connections from the 3D encoder stream aggregate low-level features along the time axis, so that they can complement the corrupted feature points. Based on our residual learning algorithm and robust loss function design, the proposed framework is ranked in the first place in the ECCV Chalearn 2018 LAP Inpainting Track2 - Video decaptioning Challenge. We hope our proposed model will become an important basic architecture for solving real-world video restoration tasks.

\paragraph{Acknowledgements}
Dahun Kim was partially supported by Global Ph.D. Fellowship Program through the National Research Foundation of Korea (NRF) funded by the Ministry of Education (NRF-2018H1A2A1062075).

{\small
\bibliographystyle{ieee}
\bibliography{egbib}
}

\end{document}